%% file: main.tex
\documentclass[lettersize,journal]{IEEEtran}
\usepackage{amsmath,amsfonts}
\usepackage{algorithmic}
\usepackage{algorithm}
\usepackage{array}
\usepackage[caption=false,font=normalsize,labelfont=sf,textfont=sf]{subfig}
\usepackage[percent]{overpic}
\usepackage{booktabs}
\usepackage{makecell}
\usepackage{textcomp}
\usepackage{stfloats}
\usepackage{url}
\usepackage{verbatim}
\usepackage{graphicx}
\usepackage{cite}
\usepackage{hyperref}
\usepackage{cleveref}
\usepackage{orcidlink}
\usepackage{balance}
\usepackage{tikz}
\usepackage{epsfig}
\usepackage{amsmath}
\usepackage{amssymb}
\usepackage{booktabs}
\usepackage{makecell}
\usepackage{multirow}
\usepackage{pifont}
\usepackage{capt-of}

\hypersetup{
    colorlinks=true,
    linkcolor=blue,
    filecolor=magenta,      
    urlcolor=cyan,
    pdfpagemode=FullScreen,
    }
\crefname{figure}{Fig.}{Figs.}
\crefname{equation}{Eq.}{Eqs.}

\begin{document}

\bstctlcite{IEEEexample:BSTcontrol}

\title{EquiGQNet: Fast Grasp Quality Evaluation
\\via Shared Equivariant Point Cloud Encoding}

\author{
Sungwon Seo$^{1}$\orcidlink{0000-0001-7862-7936},
Jaeseog Won$^{2}$\orcidlink{0009-0003-9504-5059}, 
Jiyou Shin$^{1}$\orcidlink{0009-0007-8952-1644},
Youngjin Seo$^{2}$\orcidlink{0009-0004-8247-6340}, \\ 
Hyunjun Kim$^{2}$\orcidlink{0009-0005-6260-8059},
Seokmin Yoon$^{2}$\orcidlink{0009-0006-4744-8656},
Tuan Luong$^{1}$,\orcidlink{0000-0001-5490-4337}
and Hyungpil Moon$^{1,2*}$\orcidlink{0000-0002-1091-0716}

\thanks{
This work was supported by the Robot Industrial Technology Development Program(RS-2024-00444054, Development of low-cost, high-performance visual sensor technology integrated with AI semiconductors and AI algorithms designed for robots) funded by the Ministry of Trade Industry \& Energy(MOTIE, Korea).
This research was supported by the Brain Pool program funded by the Ministry of Science and ICT through the National Research Foundation of Korea (RS-2025-15002974).
\textit{(Corresponding author: Hyungpil Moon.)}}
\thanks{
$1$: The authors are with the Faculty of Mechanical Engineering, Sungkyunkwan University, Suwon 16419, South Korea
(e-mail: ssw0536@g.skku.edu; danny384@g.skku.edu; luongtuan@g.skku.edu; hyungpil@skku.edu)
}
\thanks{
$2$: The authors are with the Faculty of Intelligent Robotics, Sungkyunkwan University, Suwon 16419, South Korea
(e-mail: jaeseogwon@g.skku.edu; soh2879@g.skku.edu; hyunjun.1104@g.skku.edu; yosmon@g.skku.edu; hyungpil@skku.edu)}
}


\maketitle
\begin{abstract}
Planning six-degree-of-freedom (6-DoF) grasps for unseen objects in cluttered tabletop scenes from a single-view depth image requires accurate and efficient evaluation of diverse grasp candidates.
Existing early-fusion methods capture local object geometry relative to each grasp candidate but repeatedly encode the scene, whereas late-fusion methods reuse a shared scene representation but may lose this grasp-relative local geometry.
We propose EquiGQNet, an efficient 6-DoF grasp quality evaluator that combines the strengths of both approaches. 
For grasp orientation, EquiGQNet replaces the early-fusion operation of rotating and re-encoding the point cloud for each grasp candidate with an SO(3)-equivariant encode-once-then-rotate scheme, yielding grasp-aligned geometric features from a shared scene encoding.
For grasp translation, Mid-level Action Fusion (MAF) injects the grasp position into intermediate features before global aggregation, retaining local geometry relative to each candidate.
We evaluate EquiGQNet in two grasp planning pipelines: Cross-Entropy Method (CEM)-based continuous grasp search and candidate ranking with a pretrained generative planner.
In simulation, EquiGQNet achieves grasping performance comparable to the early-fusion baseline and substantially outperforms late fusion on objects with complex geometry and limited graspable regions, while reducing CEM planning time from 3.31~s to 0.48~s, a 6.9$\times$ speedup over early fusion.
In real-world household-object decluttering, EquiGQNet achieves a 95.2\% grasp success rate and 230 picks per hour, versus 153 and 170 for early- and late-fusion baselines.
Code is available at \url{https://equigqnet.github.io/}.
\end{abstract}



\input{introduction}

\input{method}

\input{experiment}

\input{conclusion}

\balance
\bibliographystyle{IEEEtran}
\bibliography{reference}

\vfill

\end{document}

%% file: introduction.tex
\section{Introduction}
\label{sec:introduction}
\IEEEPARstart{R}{eliable} robotic grasping of unseen objects requires selecting an appropriate six-degree-of-freedom (6-DoF) grasp pose from limited observations.
In this work, we consider the problem of planning 6-DoF grasps with a parallel-jaw gripper for unseen objects in cluttered tabletop scenes using a point cloud obtained from a single-view depth image.
Recent learning-based grasp planning methods have achieved strong performance on objects with relatively simple and regular shapes~\cite{zhu2022sampleefficientgrasp, huang2023edgegrasp, hu2024orbitgrasp, sundermeyer2021contactgraspnet, lim2024equigraspflow, murali2025graspgen}.
However, reliable grasp planning remains challenging for objects with complex geometry~\cite{danielczuk2021exploratory}.
For such objects, successful grasps may occupy only narrow regions of the 6-DoF grasp space, making it important to explore diverse grasp candidates and evaluate them accurately.
The ability to evaluate many candidates both quickly and accurately is therefore critical to overall grasp planning performance.
Existing methods can be broadly divided into those that predict grasps at a set of spatial anchors in the scene and those that can directly evaluate arbitrary 6-DoF grasp poses.

\begin{figure}[t]
    \centering
    \includegraphics[width=1.0\linewidth]{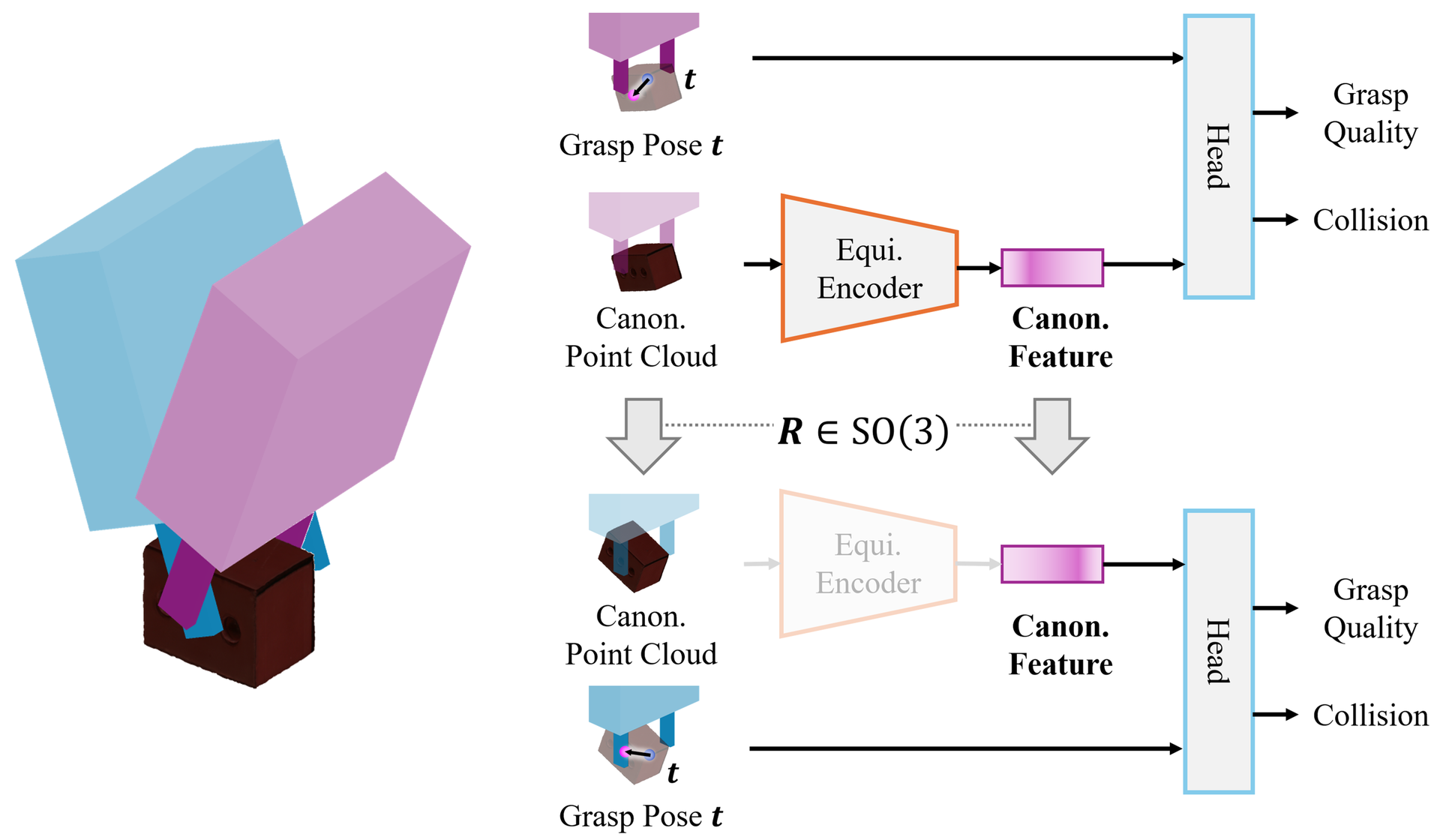}
    \caption{\textbf{Overview of EquiGQNet.}
    The left shows two different grasps of the same object.
    The two branches on the right show the point cloud rotated into each grasp frame, i.e., canonicalized, and processed by the equivariant encoder.
    By SO(3) equivariance, the relative rotation between the two canonicalized point clouds induces the same rotation between their corresponding features, allowing the feature for one grasp to be obtained from the other without an additional encoder pass.
    }
    \label{fig:overview}
    \vspace{-2mm}
\end{figure}

Scene-anchored grasp methods predict grasp poses which are associated with spatial anchors, such as pixels, points, or voxels~\cite{zhu2022sampleefficientgrasp, huang2023edgegrasp, hu2024orbitgrasp, breyer2021volumetric, jiang2021synergies}.
%
%
These methods achieve computational efficiency by encoding the scene only once and using local geometric information at the location corresponding to each candidate.
%
However, because their grasp predictions are tied to a set of spatial anchors, these methods may miss valid grasps between sampled locations.
A further limitation is that they cannot directly evaluate arbitrary candidates in the continuous 6-DoF pose space.
This restricts their applicability to grasp planning pipelines in which grasp candidates are generated independently and subsequently evaluated and ranked.

In contrast, some methods directly evaluate arbitrary 6-DoF grasp poses~\cite{murali2025graspgen, yamada2025grasp, liang2019pointnetgpd, mousavian2019graspnet, jauhri2024neugraspnet}. 
These methods can be categorized into early- and late-fusion approaches, depending on the stage at which the grasp pose is introduced.
Late-fusion methods first encode the point cloud into a shared global feature and incorporate each candidate's position and orientation only at the final prediction stage~\cite{murali2025graspgen, yamada2025grasp}. 
However, because the scene is encoded before the grasp pose is introduced, the extracted features may not explicitly capture local geometry relative to the gripper.
This limitation can be particularly consequential for irregular objects with small graspable regions, where grasp success depends strongly on local geometry.

Early-fusion methods address this issue by introducing the grasp pose before encoding the point cloud~\cite{liang2019pointnetgpd, mousavian2019graspnet, jauhri2024neugraspnet}.
For example, they may transform the scene point cloud into the coordinate frame of each grasp candidate~\cite{liang2019pointnetgpd}, or jointly encode object points with gripper reference points transformed according to the candidate pose~\cite{mousavian2019graspnet}.
This allows the encoder to process object geometry relative to the gripper's position and orientation and thus extract information such as whether the object lies between the fingers or whether nearby surfaces cause collisions.
However, every new grasp candidate requires reconstructing the input and running the entire encoder again, resulting in substantial computational cost when evaluating many candidates or performing iterative optimization.
We therefore seek a grasp evaluator that can accept arbitrary 6-DoF poses, exploit geometric information specific to each grasp as in early fusion, and still reuse a single point cloud encoding across many candidates.

\begin{figure*}[t]
    \centering
    \includegraphics[width=0.95\textwidth]{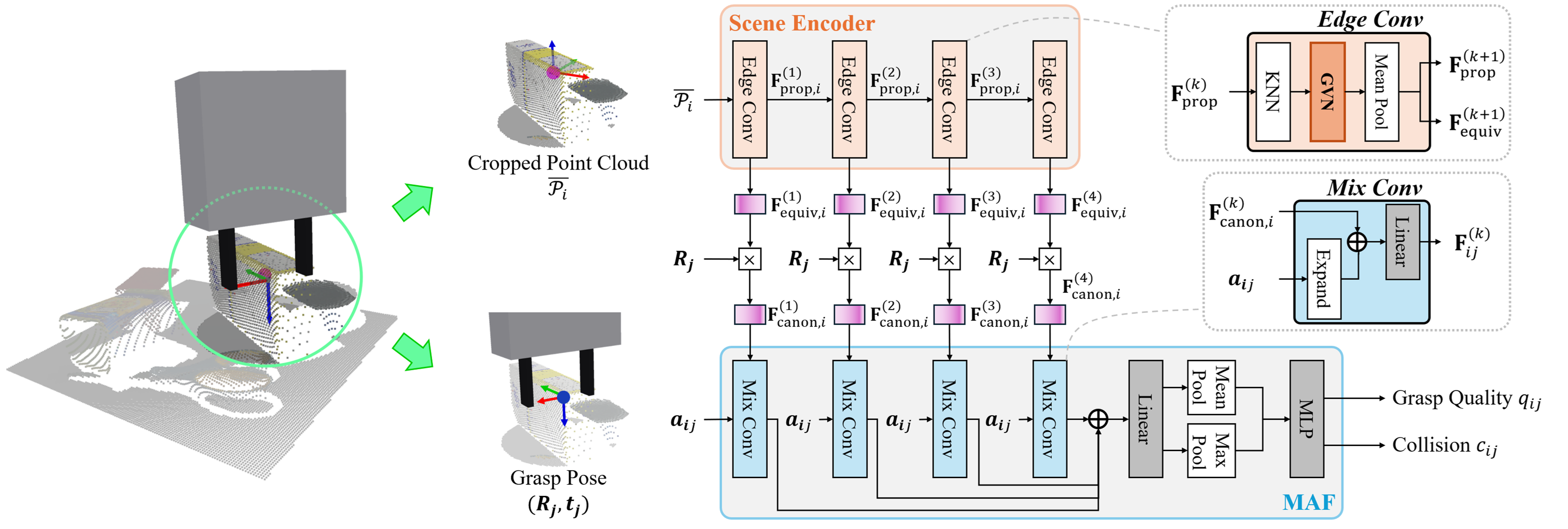}
    \caption{
    \textbf{EquiGQNet architecture.}
    A crop is encoded once into shared equivariant scene features.
    For each grasp candidate $g_j=(\mathbf{R}_j,\mathbf{t}_j)$, the shared features are aligned to the grasp coordinate frame using the inverse rotation $\mathbf{R}_j^\top$ and the aligned translation $\mathbf{a}_{ij}$ (Eqs.~\eqref{eq:feature_alignment}--\eqref{eq:aligned_translation}).
    The grasp-aligned features are then fused with $\mathbf{a}_{ij}$ through MAF for grasp quality and collision prediction.
    Here, $\times$ denotes matrix multiplication, and $\oplus$ denotes feature concatenation.
    }
    \label{fig:network}
\end{figure*}

To this end, we propose \textbf{EquiGQNet}, a 6-DoF grasp evaluation model that captures object geometry relative to each candidate grasp while reusing a single point cloud encoding across candidates.
As illustrated in Fig.~\ref{fig:overview}, EquiGQNet handles grasp rotation and translation differently.
For rotation, EquiGQNet retains the early-fusion treatment of grasp orientation by rotating the point cloud according to each candidate grasp orientation before encoding.
SO(3) equivariance allows the same grasp-oriented features to be obtained by rotating a single shared encoding, avoiding candidate-wise re-encoding.
For translation, \textbf{Mid-level Action Fusion (MAF)} incorporates the grasp position at intermediate feature layers before local geometric information is lost through global aggregation.
Together, these designs allow EquiGQNet to evaluate each grasp using object geometry relative to the gripper, as in early-fusion methods, without candidate-wise scene re-encoding.

Our main contributions are as follows:
\begin{itemize}
    \item We propose \textbf{EquiGQNet}, a 6-DoF grasp evaluator that captures object geometry relative to each grasp while reusing one scene encoding across candidates, avoiding repeated encoding in early fusion.

    \item SO(3)-equivariant feature rotation reproduces grasp-oriented scene features from a shared encoding, avoiding candidate-wise rotation and re-encoding of the point cloud, while \textbf{MAF} introduces grasp position before global aggregation to preserve local geometry around the grasp.

    \item We evaluate EquiGQNet in CEM-based continuous grasp search and generative-model-based candidate ranking.
    Experiments show reduced evaluation and planning time while maintaining strong performance, particularly on unseen objects with complex geometry and limited graspable regions.
\end{itemize}

The rest of this paper is organized as follows.
Sec.~\ref{sec:method} presents the EquiGQNet architecture, including equivariant feature alignment and Mid-level Action Fusion.
Sec.~\ref{sec:experiments} evaluates grasp performance and computational efficiency through simulation, ablation studies, and real-world experiments.
Sec.~\ref{sec:conclusion} concludes the paper.

%% file: method.tex
\section{METHOD}
\label{sec:method}
As illustrated in Fig.~\ref{fig:network}, EquiGQNet encodes a local point-cloud crop once into shared $\mathrm{SO}(3)$-equivariant features.
For each grasp candidate, these features are rotated according to its orientation and fused with its position through MAF before global aggregation.

Let an observed scene be represented by a point cloud
$\mathcal{P}=\{\mathbf{p}_n\}_{n=1}^{N}$, where $\mathbf{p}_n\in\mathbb{R}^{3}$.
A 6-DoF grasp pose is represented as
$g=(\mathbf{R},\mathbf{t})\in\mathrm{SE}(3)$,
where $\mathbf{R}\in\mathrm{SO}(3)$ and $\mathbf{t}\in\mathbb{R}^{3}$ denote the gripper orientation and position, respectively.
Our goal is to predict grasp robustness $q\in[0,1]$, which reflects grasp success under perturbations, while separately predicting collision probability.
We adopt the column vector convention for 3-D rotations,
$\mathbf{v}'=\mathbf{R}\mathbf{v}$, while storing vector-valued feature channels as rows.
Accordingly, $\mathbf{V}\in\mathbb{R}^{C\times 3}$ transforms as
$\mathbf{V}\mapsto\mathbf{V}\mathbf{R}^{\top}$.

\subsection{SO(3)-Equivariant Scene Encoder}
\label{sec:equivariant_encoder}
Our scene encoder represents each object’s local geometry with $\mathrm{SO}(3)$-equivariant vector features by encoding an object-centered crop once and reusing its features across grasp candidates.

\subsubsection{Point Cloud Crop}
Rather than processing the entire observed scene, EquiGQNet constructs a local point cloud around each crop center.
Let $\mathbf{c}_i\in\mathbb{R}^{3}$ denote the center of the $i$-th crop, and let $r_c$ denote the crop radius.
The network input is defined as
\begin{equation}
\bar{\mathcal{P}}_i
=
\left\{
\mathbf{p}_n-\mathbf{c}_i
\;\middle|\;
\mathbf{p}_n\in\mathcal{P},
\;
\left\|\mathbf{p}_n-\mathbf{c}_i\right\|_2\le r_c
\right\}.
\end{equation}
Here, $N_i = |\bar{\mathcal{P}}_i|$ denotes the number of points in the $i$-th crop, such that $\bar{\mathcal{P}}_i \in \mathbb{R}^{N_i \times 3}$.
This crop centered local representation excludes distant scene geometry, while retaining the spatial context needed to evaluate nearby grasp candidates.

\subsubsection{Gated Vector Neurons}
Vector Neurons (VN)~\cite{deng2021vectorneurons} represent each feature channel as a 3-D vector to preserve rotational equivariance.
Inspired by GVP-style vector gating~\cite{jing2021learning}, our lightweight Gated Vector Neurons (GVN) replace VN-ReLU’s sign-based half-space truncation~\cite{deng2021vectorneurons} with smooth multiplicative gates from batch-normalized channel-wise inner products and a learned sigmoid.

Given $\mathbf{V}\in\mathbb{R}^{C\times3}$, GVN computes
\begin{equation}
\begin{gathered}
\mathbf{K}=W_K\mathbf{V},
\qquad
\mathbf{Q}=W_Q\mathbf{V},\\
\boldsymbol{g}
=
\sigma\!\left(
W_g\,\operatorname{BN}
\bigl(\operatorname{diag}(\mathbf{K}\mathbf{Q}^{\top})\bigr)
\right),\\
\operatorname{GVN}(\mathbf{V})
=
\boldsymbol{g}\odot\mathbf{K}.
\end{gathered}
\end{equation}
where 
$W_K,W_Q\in\mathbb{R}^{C'\times C}$ and
$W_g\in\mathbb{R}^{C'\times C'}$ are learned weight matrices,
$\operatorname{BN}(\cdot)$ denotes batch normalization, and
$\sigma(\cdot)$ is the sigmoid function.
The projected vector features are
$\mathbf{K},\mathbf{Q}\in\mathbb{R}^{C'\times3}$, while
$\operatorname{diag}(\mathbf{K}\mathbf{Q}^{\top}) \in \mathbb{R}^{C'}$
contains the inner products between their corresponding channels.
The operator $\odot$ denotes channel-wise scalar multiplication over the 3-D vector dimension.
Thus, GVN preserves $\mathrm{SO}(3)$ equivariance while providing a lightweight gated nonlinearity:
\begin{equation}
\operatorname{GVN}
\left(
\mathbf{V}\mathbf{R}^{\top}
\right)
=
\operatorname{GVN}(\mathbf{V})
\mathbf{R}^{\top},
\qquad
\forall \mathbf{R}\in \mathrm{SO}(3).
\end{equation}

\subsubsection{Equivariant EdgeConv Encoder}
We encode $\bar{\mathcal{P}}_i$ using an \emph{Equivariant EdgeConv Encoder} adapted from the dynamic graph architecture of VN-GCNN~\cite{deng2021vectorneurons,wang2019dynamic}.
We use $\bar{\mathcal{P}}_i$ directly as the initial vector features, $\mathbf{F}_{\mathrm{prop},i}^{(0)}=\bar{\mathcal{P}}_i$.
The encoder consists of $L$ successive \emph{EdgeConv} levels, indexed by $l\in\{1,\ldots,L\}$, where a $k$-nearest-neighbor graph is dynamically recomputed from the propagated vector features at each level.

Omitting crop index $i$, let $\mathbf{f}_n$ denote the vector feature of point $n$.
For each point $n$ and its neighbor $m\in\mathcal{N}_{n}^{(l)}$, we form an edge feature, apply GVN, and mean-pool over the neighborhood:
\begin{equation}
\begin{aligned}
\Delta\mathbf{f}_{nm}^{(l-1)}
&=
\mathbf{f}_{\mathrm{prop},m}^{(l-1)}
-
\mathbf{f}_{\mathrm{prop},n}^{(l-1)},
\\
\mathbf{h}_{nm}^{(l)}
&=
\operatorname{GVN}^{(l)}
\left(
\left[
\Delta\mathbf{f}_{nm}^{(l-1)},
\mathbf{f}_{\mathrm{prop},n}^{(l-1)}
\right]
\right),
\\
\left[
\mathbf{f}_{\mathrm{equiv},n}^{(l)},
\mathbf{f}_{\mathrm{prop},n}^{(l)}
\right]
&=
\operatorname{MeanPool}_{m\in\mathcal{N}_{n}^{(l)}}
\left(
\mathbf{h}_{nm}^{(l)}
\right).
\end{aligned}
\label{eq:equivariant_edgeconv}
\end{equation}
Here, $\mathcal{N}_{n}^{(l)}$ denotes the dynamically constructed $k$-nearest-neighbor set of point $n$ at encoder level $l$.

Restoring the crop index $i$, the resulting features are $\mathbf{F}_{\mathrm{equiv},i}^{(l)} \in \mathbb{R}^{N_i \times C_{\mathrm{equiv}}^{(l)} \times 3}$ and $\mathbf{F}_{\mathrm{prop},i}^{(l)} \in \mathbb{R}^{N_i \times C_{\mathrm{prop}}^{(l)} \times 3}$, where $C_{\mathrm{equiv}}^{(l)}$ and $C_{\mathrm{prop}}^{(l)}$ denote the respective numbers of vector feature channels.
The equivariant feature $\mathbf{F}_{\mathrm{equiv},i}^{(l)}$ is retained for alignment and fusion for each grasp candidate, whereas $\mathbf{F}_{\mathrm{prop},i}^{(l)}$ is propagated to the next \textit{EdgeConv} layer.

Because all EdgeConv operations preserve rotational equivariance, both outputs remain $\mathrm{SO}(3)$-equivariant:
\begin{equation}
\begin{aligned}
\operatorname{EdgeConv}^{(l)}
\left(
\mathbf{F}_{\mathrm{prop},i}^{(l-1)}
\mathbf{R}^{\top}
\right)
=\operatorname{EdgeConv}^{(l)}
\left(
\mathbf{F}_{\mathrm{prop},i}^{(l-1)}
\right)
\mathbf{R}^{\top},
\end{aligned}
\label{eq:edgeconv_equivariance}
\end{equation}
$\forall \mathbf{R}\in\mathrm{SO}(3)$. Both features are independent of the grasp candidate and are therefore computed only once for each point cloud crop.

\subsubsection{Grasp Frame Feature Alignment}
For grasp $g_j=(\mathbf{R}_j,\mathbf{t}_j)$, its pose relative to the crop center is $(\mathbf{R}_j,\mathbf{t}_j-\mathbf{c}_i)$.
Expressing the scene in the grasp coordinate frame therefore corresponds to applying its inverse transform, $(\mathbf{R}_j^\top,-\mathbf{R}_j^\top(\mathbf{t}_j-\mathbf{c}_i))$.
Accordingly, the cached equivariant feature at level $l$ is aligned with the grasp orientation as
\begin{equation}
\mathbf{F}_{\mathrm{canon},ij}^{(l)}
=
\mathbf{F}_{\mathrm{equiv},i}^{(l)}
\mathbf{R}_j.
\label{eq:feature_alignment}
\end{equation}
By Eq.~\eqref{eq:edgeconv_equivariance}, under our row-vector convention, right multiplication by $\mathbf{R}_j$ yields the same feature as rotating the crop into the grasp frame by $\mathbf{R}_j^\top$ and re-encoding it.
Thus, rotating the cached feature is equivalent to first transforming the point cloud crop into the grasp-oriented frame and then re-encoding it.
The crop origin expressed in the grasp frame is
\begin{equation}
\mathbf{a}_{ij}
=
-\mathbf{R}_j^\top
\left(
\mathbf{t}_j-\mathbf{c}_i
\right).
\label{eq:aligned_translation}
\end{equation}

\subsection{Mid-level Action Fusion}
\label{sec:maf}

After orientation alignment, grasp-dependent translation $\mathbf{a}_{ij}$ must interact with local geometry; appending it only after global aggregation, as in late fusion, discards per-point spatial information before this interaction.
We propose \emph{Mid-level Action Fusion} (MAF), which injects the grasp translation into the canonical scene features at multiple encoder levels before global aggregation.
At level $l$, $\mathbf{a}_{ij}$ is broadcast across crop points and fused point-wise with the canonical feature using MixConv:
\begin{equation}
\mathbf{F}_{ij}^{(l)}
=
\operatorname{MixConv}^{(l)}
\left(
\left[
\mathbf{F}_{\mathrm{canon},ij}^{(l)};
\;
\mathbf{1}_{N_i}\mathbf{a}_{ij}^{\top}
\right]
\right),
\label{eq:mixconv}
\end{equation}
where
$\mathbf{F}_{\mathrm{canon},ij}^{(l)}
\in \mathbb{R}^{N_i\times C_{\mathrm{equiv}}^{(l)}\times 3}$
and $\mathbf{1}_{N_i}\mathbf{a}_{ij}^{\top}\in\mathbb{R}^{N_i\times3}$
broadcasts $\mathbf{a}_{ij}$ over all points in crop $i$ and is treated as an additional 3-D vector channel, yielding an
$N_i\times(C_{\mathrm{equiv}}^{(l)}+1)\times3$ input to \textit{MixConv}.
The aligned vector components are flattened along the channel dimension before \textit{MixConv}.
The resulting $\mathbf{F}_{ij}^{(l)}$ thus contains information from both the geometry of crop $i$ and the translation of candidate $j$.

The fused features from all encoder levels are concatenated along the channel dimension and projected using a point-wise linear layer, followed by max--mean aggregation:
\begin{equation}
\begin{aligned}
\mathbf{H}_{ij}
&=
\operatorname{Linear}
\left(
[\mathbf{F}_{ij}^{(l)}]_{l=1}^{L}
\right),\\
\mathbf{h}^{\mathrm{global}}_{ij}
&=
\left[
\max_n \mathbf{H}_{ij,n};
\frac{1}{N_i}\sum_{n=1}^{N_i}\mathbf{H}_{ij,n}
\right].
\end{aligned}
\label{eq:maf_aggregation}
\end{equation}
Fusing the translation before aggregation preserves spatial information relative to the grasp position across multiple receptive-field scales.
The pooled feature is then passed through a lightweight MLP to predict grasp robustness $\hat{q}_{ij}$ and collision probability $\hat{c}_{ij}$.

%% file: experiment.tex
\section{Experiments}
\label{sec:experiments}

\subsection{Training Data and Implementation Details}
\label{sec:exp_training}

\subsubsection{Synthetic Training Data}
We build a synthetic grasp dataset from IPA-3D1K~\cite{lindermayr2023ipa}, which mainly contains regular retail objects, and the more diverse and complex Thingi10K~\cite{zhou2016thingi10k}.
For training, we randomly sample 500 objects from IPA-3D1K and Thingi10K with a 6:4 ratio.
For validation, we use an additional disjoint set of 100 objects sampled with the same ratio.
Following the object-to-gripper scale convention commonly used in grasping benchmarks~\cite{morrison2020egad}, meshes whose middle spatial extent exceeds the maximum gripper opening of $80\, \mathrm{mm}$ are uniformly scaled such that their middle extent falls between $0.5$ and $0.8$ times the gripper opening.

We generate cluttered scenes in PyBullet~\cite{coumans2021pybullet} by dropping between 1 and 12 objects into the workspace and allowing them to settle under physics.
In total, 5,000 training scenes and 1,000 validation scenes are generated from the corresponding object splits.
For each object instance in a scene, we sample 100 grasp candidates using the antipodal and approach-based sampling schemes of \cite{eppner2019billion}.
We construct the candidate set with 50\% collision-free antipodal grasps, 25\% antipodal grasps with collisions allowed, and 25\% approach-based grasps with collisions allowed.
This mixture exposes the evaluator to diverse grasp configurations, including stable, unstable, and geometrically infeasible candidates.

We define grasp robustness as the empirical success rate over 25 physics trials, with success defined as lifting the object by at least 0.15 m, under independent Gaussian perturbations to translation ($\sigma=2.5\,\mathrm{mm}$), orientation ($\sigma=1^\circ$), object density ($\sigma=200\,\mathrm{kg/m^3}$, clipped to $[500,2000]\,\mathrm{kg/m^3}$), and lateral friction ($\sigma=0.1$, clipped to $[0.1,1.0]$), with spinning friction varied proportionally to lateral friction; grasps in collision are assigned a robustness of zero; cf. \cite{weisz2012pose,mahler2017dexnet2}.
The resulting training and validation sets contain approximately 2.5 million and 0.6 million grasps, respectively, with roughly 20\% positive and 80\% negative samples in both sets using a robustness threshold of 0.5.

For rendering, the camera is placed at a nominal distance of $1.0\,\mathrm{m}$ from the scene, with uniform distance perturbation of $\pm0.25\,\mathrm{m}$.
Its nominal viewing direction is tilted by $20^\circ$ from the downward direction, with an additional uniform perturbation of $\pm15^\circ$.
The resulting depth observations are converted to point clouds and used as network inputs.

\begin{figure}[t]
    \centering
    \includegraphics[width=1.0\linewidth]{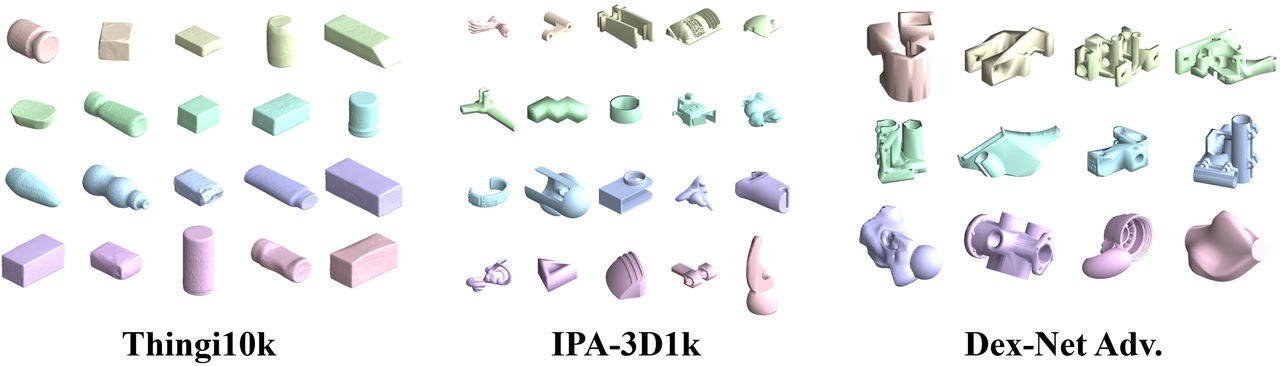}
    \caption{
    Simulation object sets used for grasp evaluation: IPA-3D1K~\cite{lindermayr2023ipa}, Thingi10K~\cite{zhou2016thingi10k}, and scaled Dex-Net adversarial objects~\cite{mahler2017dexnet2}.
    }
    \label{fig:sim_objects}
\end{figure}
\input{tables/main_results_bundle}

\subsubsection{Training Details}
Unless otherwise specified, we train all models for five epochs using AdamW with an initial learning rate of $10^{-3}$, weight decay of $10^{-2}$, and learning-rate decay of $0.95$ every $0.2$ epoch.
EquiGQNet uses four equivariant EdgeConv levels with $k=20$, one retained equivariant channel per level, propagated widths $[10,10,21]$, and MAF widths $[30,30,63,126]$, followed by a 512-D projection and a $512$--$256$ MLP head.
We use a crop radius of $r_c=0.1\,\mathrm{m}$ and resample each crop to 1024 points using Farthest Point Sampling (FPS) for larger crops and sampling with replacement for smaller ones.
Crop centers are object-mask centroids perturbed by uniform noise of up to $0.02\,\mathrm{m}$; after cropping, we apply zero-mean Gaussian translation augmentation with $\sigma=0.005\,\mathrm{m}$.
We optimize $\mathcal{L}=\mathcal{L}_q+\lambda\mathcal{L}_c$ with $\lambda=1$, where $\mathcal{L}_q$ and $\mathcal{L}_c$ are BCE losses for the robustness score $q\in[0,1]$ as a soft target and the collision label $c\in\{0,1\}$ indicating gripper--scene intersection, respectively.
No class reweighting or resampling is used.
All training and simulation experiments use an AMD Ryzen 7 7700 CPU and an NVIDIA RTX 3090 Ti GPU.


\subsection{Simulation Evaluation Protocol}
\label{sec:simulation_protocol}

We evaluate all grasps in PyBullet on three geometrically distinct object sets, shown in Fig.~\ref{fig:sim_objects}. We randomly select 20 IPA-3D1K objects and curate 20 geometrically challenging Thingi10K objects, excluding bulky primitive-like shapes that admit a wide range of valid grasp placements. All objects are disjoint from the training and validation splits. We additionally evaluate on the Dex-Net adversarial objects
\cite{mahler2017dexnet2}.
These objects were originally evaluated using an ABB YuMi gripper with an approximately $50\,\mathrm{mm}$ opening.
Because our gripper has an $80\,\mathrm{mm}$ maximum opening, we uniformly scale the objects by a factor of $1.6$ to preserve the relative object-to-gripper scale \cite{morrison2020egad}.

We consider both isolated-object and cluttered settings.
In the isolated setting, each test object is placed with a random pose and evaluated for 100 grasp trials.
For IPA-3D1K and Thingi10K, we additionally generate 100 cluttered scenes per dataset, each containing five objects.
Objects that leave the valid workspace are removed, and a cluttered episode is terminated after two consecutive failed grasp attempts, following the protocol used by OrbitGrasp \cite{hu2024orbitgrasp}.
We report grasp success rate (GSR) in both settings and declutter rate (DR) for cluttered scenes.
The Dex-Net adversarial set is evaluated only in the isolated setting due to its larger object scale, which makes consistent cluttered-scene construction difficult within the fixed workspace.

\subsection{Grasp Planning Pipelines}
\label{sec:grasp_plannig_pipelines}
A grasp evaluator does not itself generate candidate poses, so we evaluate arbitrary-pose evaluators under two complementary planning strategies.
Although the evaluator accepts arbitrary continuous 6-DoF poses, both strategies impose a $45^\circ$ approach-angle constraint, as detailed below. 
For evaluators that predict collision probability, candidates are ranked by $\hat{q}(1-\hat{c})$.
To characterize the computational cost of processing multiple local regions, we denote by $N_q$ the number of point cloud crops (local scene queries) evaluated per observation, distinct from the number of grasp candidates scored within each crop.

\smallskip
\noindent\textbf{Continuous refinement with CEM:} 
We consider continuous refinement using CEM initialized with image-based antipodal grasp sampling \cite{mahler2017dexnet2}.
For each object detected by Unseen Object Instance Segmentation (UOIS)~\cite{xie2021unseen}, we construct a local point cloud crop centered at the centroid of its masked point cloud and sample 256 initial grasp candidates per crop.
Because the antipodal sampler produces top-down grasps, we perturb their orientations using local 3-D axis--angle noise with a $15^\circ$ per-axis standard deviation.
CEM operates on a standardized 9-D pose representation consisting of translation and the first two columns of the rotation matrix, with sampled rotations projected back to $SO(3)$ by Gram--Schmidt orthonormalization.
We run 10 CEM iterations with 128 candidates per iteration and retain the top 25\% as elites, resulting in up to 1,536 grasp evaluations per local query.
Candidates violating the $45^\circ$ approach-angle constraint or lying outside the crop radius $r_c$ are assigned zero quality.

\smallskip
\noindent\textbf{Samples from a pretrained generative model:}
We evaluate the learned evaluators as scorers for a generative grasp planner.
We use the pretrained GraspGen generator \cite{murali2025graspgen}, which takes an object-segmented point cloud as input, to produce 1,024 candidate grasps per object.
In simulation, the generator input is constructed using ground-truth instance masks provided by the simulator.
From these 1,024 candidates, those exceeding $45^\circ$ tilt constraint are removed, and the remaining grasps are scored by the evaluator under test.
The highest-scoring candidate is selected for execution.

\begin{figure}[t]
    \centering

    \begin{minipage}[t]{0.05\linewidth}
        \vspace{0pt}
        (a)
    \end{minipage}%
    \begin{minipage}[t]{0.95\linewidth}
        \vspace{0pt}
        \includegraphics[width=\linewidth, height=3cm]
        {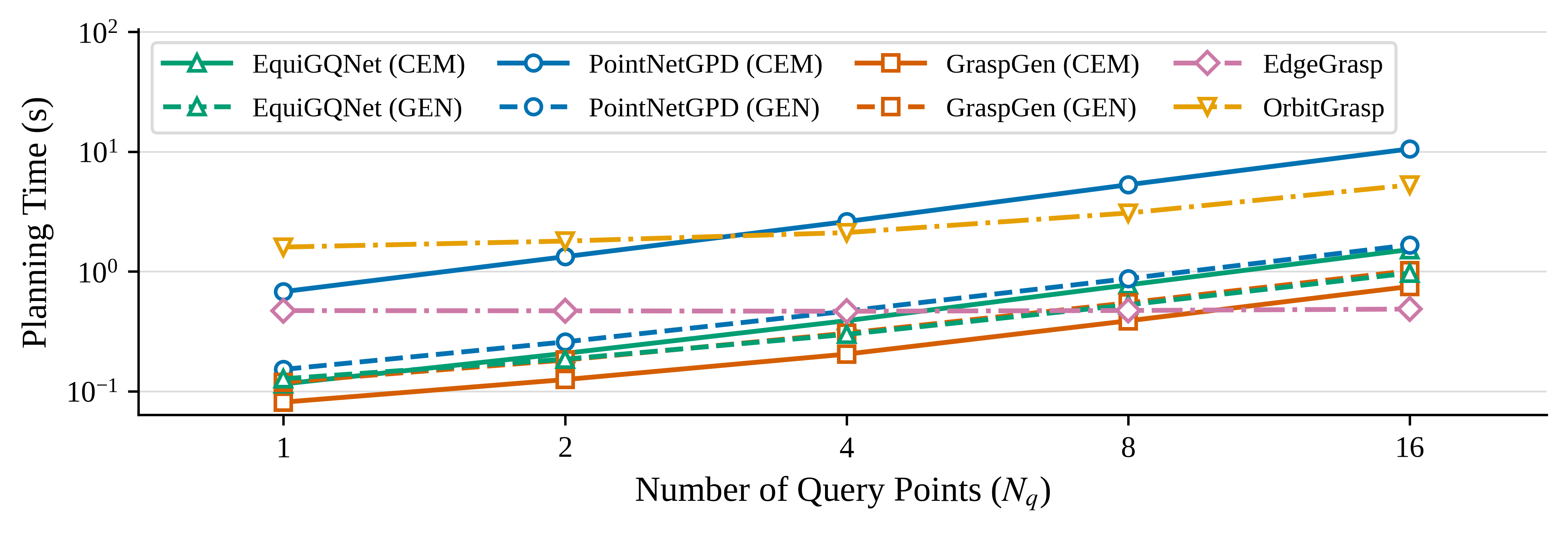}
    \end{minipage}

    \begin{minipage}[t]{0.05\linewidth}
        \vspace{0pt}
        (b)
    \end{minipage}%
    \begin{minipage}[t]{0.95\linewidth}
        \vspace{0pt}
        \includegraphics[width=\linewidth, height=3cm]
        {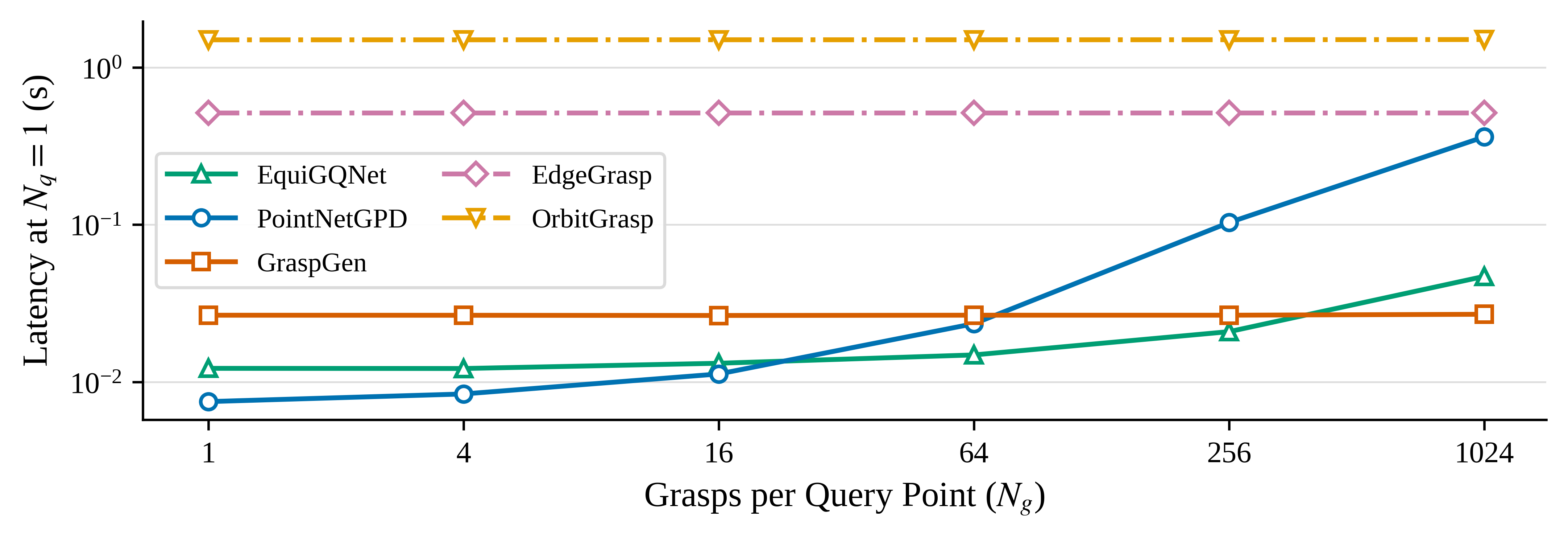}
    \end{minipage}

    \vspace{-1mm}
    
    \caption{
    Planning latency. (a) Planning time versus the number of query points $N_q$. (b) Latency per query point versus the number of grasps $N_g$ scored at a single query point, including per-query preprocessing.
    }
    \label{fig:planning_time}
\end{figure}

\subsection{Comparison with Baseline Methods in Simulation}
\smallskip
\noindent\textbf{Baselines:}
We compare EquiGQNet with both arbitrary-pose grasp evaluators and scene-anchored grasp methods.
PointNetGPD~\cite{liang2019pointnetgpd} serves as an early-fusion baseline, explicitly transforming scene geometry into each candidate grasp frame before encoding.
We train PointNetGPD on our dataset while retaining its original gripper-region input convention and grasp quality prediction formulation.
As a late-fusion baseline, we train the GraspGen discriminator~\cite{murali2025graspgen} using the same object-centered crop pipeline and training data as EquiGQNet, with an initial learning rate of $10^{-4}$ for training stability; the discriminator predicts both grasp quality and collision probability.

For a broader comparison of grasping performance, we additionally include EdgeGrasp~\cite{huang2023edgegrasp} and OrbitGrasp~\cite{hu2024orbitgrasp} using their publicly available pretrained models.
As scene-anchored grasp methods, they evaluate grasps tied to sampled scene locations rather than arbitrary 6-DoF poses and are not retrained on our dataset; their results therefore serve as broader end-to-end pipeline references rather than controlled evaluator comparisons.
For OrbitGrasp, we use the same UOIS-based object-centered queries as in the CEM pipeline described in Sec.~\ref{sec:grasp_plannig_pipelines}.
EdgeGrasp instead operates without object-instance segmentation and evaluates local neighborhoods around 32 approach points selected by farthest-point sampling (FPS), corresponding to $N_q=32$.

\input{tables/validation_accuracy}

\noindent\textbf{Results:}
The simulation results across the different planning strategies are summarized in Table~\ref{tab:baselines}.
EquiGQNet consistently outperforms the late-fusion GraspGen discriminator on objects with challenging geometry, while achieving grasping performance comparable to the early-fusion PointNetGPD.
At the same time, EquiGQNet avoids the high cost of candidate-wise scene encoding: at $N_q=5$, it reduces CEM planning time from 3.308\,s for PointNetGPD to 0.478\,s, corresponding to a $6.9\times$ speedup.

The computational scaling with the number of local queries $N_q$ and grasps per query $N_g$ is shown in Fig.~\ref{fig:planning_time}.
For a fair comparison between the CEM-based and generative pipelines, we report planning latency without UOIS-based instance segmentation, which takes 54.7\,ms on average.
EquiGQNet scales more favorably with $N_q$ than PointNetGPD by reusing the scene representation (Fig.~\ref{fig:planning_time}(a)), and the gap increases with $N_g$ (Fig.~\ref{fig:planning_time}(b)).
EdgeGrasp and OrbitGrasp remain nearly flat because preprocessing accounts for over 90\% of their per-query latency.

On the Thingi10K cluttered setting, increasing CEM iterations from 3 to 10 improves GSR/DR by 1.6/3.4 pp at an additional 0.208\,s, while increasing GraspGen candidates from 256 to 1024 improves GSR/DR by 6.1/6.8 pp at an additional 0.175\,s.
We therefore use 10 CEM iterations and 1024 candidates in the main experiments.

\subsection{Ablation Study in Simulation}
We ablate how grasp rotation and translation interact with scene features, as summarized in Table~\ref{tab:ablation}.
All variants use the same CEM-based refinement pipeline.
Here, \emph{global} denotes late fusion after global pooling, \emph{canon.} denotes early-fusion scene encoding in the grasp frame, and MAF fuses the corresponding grasp pose component with scene features before aggregation.
Variant (h) uses the same standardized 9-D pose representation as CEM.
Following VN~\cite{deng2021vectorneurons}, the DGCNN variants use the same network depth and number of channels per layer as their equivariant counterparts, with $C$ scalar channels corresponding to $C$ vector channels ($C\times3$).

\begin{figure}[t]
    \centering
    \includegraphics[width=1.0\linewidth]{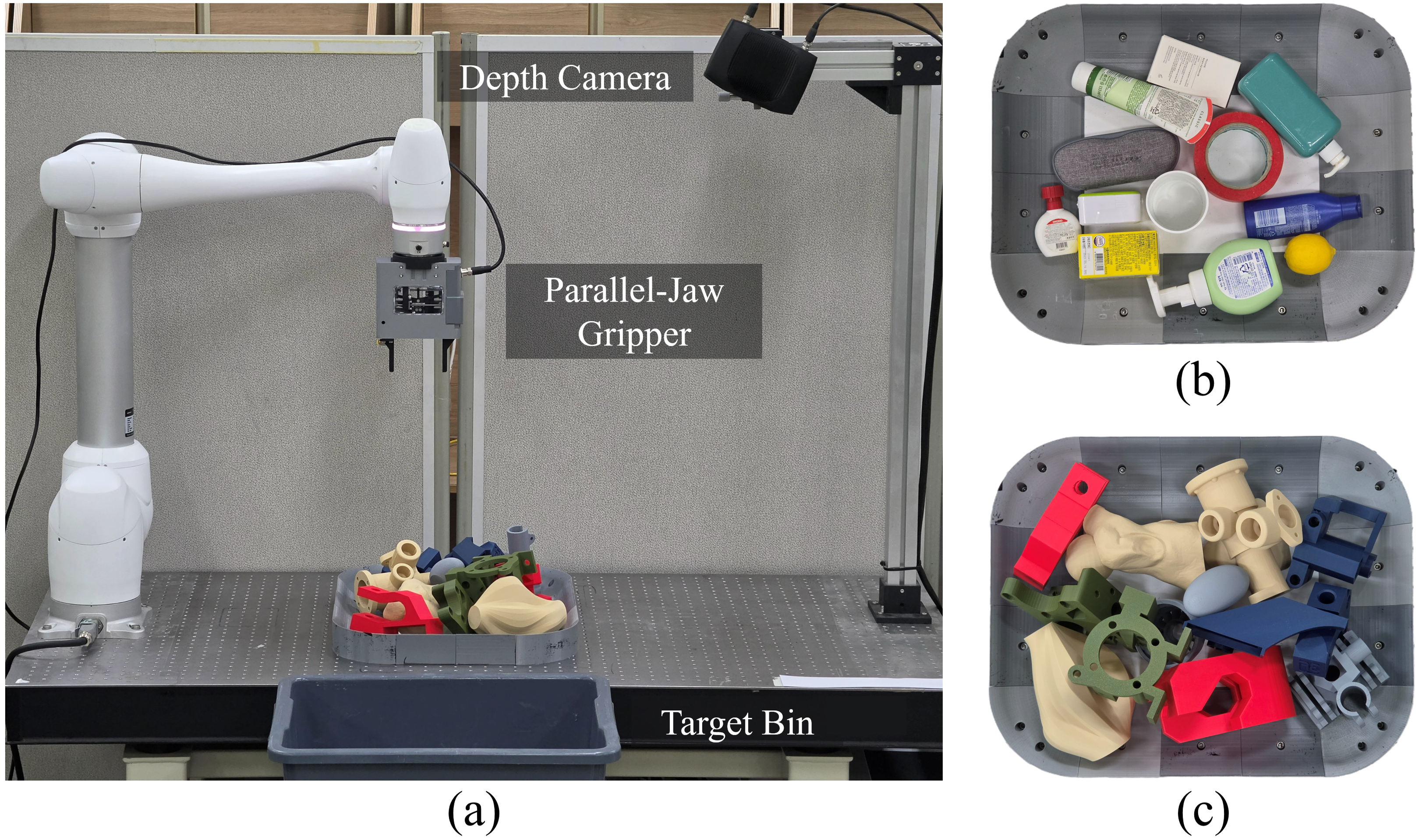}
    \caption{
    Real-world experimental setup and cluttered test objects.
    (a) Robot and camera setup.
    (b) Twelve household objects.
    (c) Twelve 1.6$\times$-scaled Dex-Net adversarial objects~\cite{mahler2017dexnet2}.
    }
    \label{fig:real_exp_setup}
\end{figure}

\smallskip
\noindent\textbf{Effect of Early Scene--Grasp Fusion:}
Variants (a)--(d) progressively introduce scene--grasp interaction earlier in the network.
While global late fusion in (a) is efficient, it performs poorly on challenging geometries; candidate-wise canonicalization in (b)--(d) substantially improves performance, but requires repeated scene encoding and approximately 7\,s of planning time.
Within this setting, the translation fusion stage is particularly important.
With identical rotation canonicalization and DGCNN encoders, replacing global translation fusion in (b) with MAF in (c) improves Thingi10K cluttered GSR/DR from 63.2/63.8\% to 85.6/89.0\%, with essentially unchanged planning time.
Moreover, (c) performs comparably to canonicalizing both rotation and translation in (d), showing that pre-aggregation translation fusion recovers most of the benefit of full pose canonicalization without the additional translation-specific scene re-encoding.
These results show that early scene--grasp interaction is beneficial, and that MAF provides an effective way to introduce translation-dependent geometric interaction before global aggregation.

\smallskip
\noindent\textbf{Efficient Pose Conditioning with Shared Encoding:}
Although (c) achieves strong performance, candidate-wise canonicalization requires re-encoding the scene for every grasp, resulting in 7.222\,s of planning time.
In contrast, equivariant feature reuse in (e) and EquiGQNet enables candidate-specific rotation alignment from a shared scene encoding, reducing planning time to 0.481 and 0.478\,s, respectively.
Moreover, replacing VN-DGCNN with GVN-DGCNN maintains comparable performance and inference cost while increasing training throughput from 5.45 to 7.45 iter/s.

An alternative way to retain shared scene encoding is (h), which replaces the explicit rotation canonicalization of (c) with rotation MAF.
However, (h) suffers a substantial performance drop, indicating that explicit alignment is more effective for rotation conditioning.
Likewise, replacing the translation MAF of EquiGQNet with global fusion in (g) leads to a large performance degradation.
Notably, the fully global GVN-DGCNN variant (f) also performs substantially worse than its DGCNN counterpart (a), while adding rotation alignment in (g) improves over (f) but still yields relatively weak performance.
These results suggest that simply applying global late fusion to shared equivariant features is insufficient, and that grasp-dependent interaction before global aggregation is important.
Overall, effective shared encoding requires explicit rotation alignment together with translation fusion before global aggregation.

\input{tables/physical_experiments}

\smallskip
\noindent\textbf{Grasp Quality Prediction Analysis:}
We analyze the grasp quality prediction quantitatively on the validation set and qualitatively using grasp robustness maps.
EquiGQNet closely matches the strongest candidate-wise canonicalization variants while retaining shared scene encoding, performs comparably to PointNetGPD, and consistently outperforms the GraspGen discriminator (Table~\ref{tab:validation_accuracy}).
These results show that it retains most of the predictive benefit of candidate-wise canonicalization while substantially reducing planning cost by reusing the scene encoding.

To examine how fusion strategies affect the spatial distribution of predicted grasp quality, we fix the grasp orientation and height and evaluate quality over a dense grid of planar translations.
The ground-truth and predicted quality maps for an IPA-3D1K object and a Dex-Net adversarial object are compared in Fig.~\ref{fig:grasp_robustness_map}.
Variants (a), (b), (f), and (g), as well as the GraspGen discriminator, produce relatively diffuse quality maps, particularly on the adversarial geometry.
In contrast, methods that incorporate both grasp rotation and translation before global aggregation produce more localized high-quality regions that better match the ground truth.
This difference is particularly important for objects with narrow graspable regions, where diffuse predictions can create spurious maxima and lead to incorrect grasp selection.

To examine whether GraspGen's on-generator training alleviates the diffuse predictions, we additionally visualize the official pretrained discriminator (GraspGen$^*$).
Although it produces more localized predictions than our retrained baseline on the simpler object, both remain diffuse on the adversarial geometry, suggesting that late grasp-pose incorporation remains limiting even with on-generator training.

\smallskip
\noindent\textbf{Effect of Collision Probability in Grasp Scoring:}
Using $\hat{q}$ alone instead of $\hat{q}(1-\hat{c})$ changes GSR by at most 0.5 percentage points across all simulation settings, indicating only a marginal benefit from explicitly using collision probability in grasp scoring, as collision is already reflected in the robustness target.

\subsection{Real-World Experiments}
\label{sec:real_experiments}
We evaluate the grasping pipeline using a Doosan Robotics M1013 with a custom $80\,\mathrm{mm}$ parallel-jaw gripper and a fixed eye-to-hand Zivid Two camera (Fig.~\ref{fig:real_exp_setup}).
All methods in Table~\ref{tab:real_experiment} use the same CEM-based continuous refinement as in simulation, differing only in the grasp evaluator.
Each run starts with 12 objects and terminates after two consecutive failed attempts, with collision-induced emergency stops counted as failures.
We perform ten decluttering runs per method and object set.
We report GSR, DR, and mean picks per hour (MPPH), measured from image acquisition through pick-and-place completion and return to the ready pose.
EquiGQNet achieves grasp success comparable to PointNetGPD while attaining the highest MPPH on both object sets.
The GraspGen discriminator shows a larger performance drop than in simulation, particularly on the Dex-Net adversarial objects.

%% file: tables/main_results_bundle.tex

\begin{figure*}[!t]
\centering

\begin{minipage}{\textwidth}
\centering
\captionof{table}{COMPARISON OF GRASP EVALUATION METHODS ACROSS DIFFERENT SEARCH STRATEGIES}
\label{tab:baselines}
\resizebox{\textwidth}{!}{%
\begin{tabular}{llc ccc ccc c c}
\toprule
\multirow{3}{*}{Search Strategy}
&
\multirow{3}{*}{Evaluation Method}
&
\multirow{3}{*}{\makecell{Shared\\Scene\\Encoding}}
&
\multicolumn{3}{c}{IPA-3D1K}
&
\multicolumn{3}{c}{Thingi10K}
&
\multicolumn{1}{c}{Dex-Net Adv.}
&
\multirow{3}{*}{\makecell{Planning\\Time (s)\\($N_q=5$)}}
\\
\cmidrule(lr){4-6}
\cmidrule(lr){7-9}
\cmidrule(lr){10-10}
&
&
&
\multicolumn{1}{c}{\makecell{Isolated\\Object}}
&
\multicolumn{2}{c}{Cluttered}
&
\multicolumn{1}{c}{\makecell{Isolated\\Object}}
&
\multicolumn{2}{c}{Cluttered}
&
\multicolumn{1}{c}{\makecell{Isolated\\Object}}
&
\\
\cmidrule(lr){4-4}
\cmidrule(lr){5-6}
\cmidrule(lr){7-7}
\cmidrule(lr){8-9}
\cmidrule(lr){10-10}
&
&
&
GSR (\%)
&
GSR (\%)
&
DR (\%)
&
GSR (\%)
&
GSR (\%)
&
DR (\%)
&
GSR (\%)
&
\\

\midrule

\multirow{3}{*}{Continuous refinement with CEM}
& EquiGQNet
& $\checkmark$
& $\mathbf{100 \pm 0.1}$
& $\mathbf{99.8 \pm 0.2}$
& $\mathbf{95.2 \pm 0.9}$
& $\underline{89.5 \pm 2.8}$
& $\underline{86.0 \pm 1.6}$
& $\underline{92.2 \pm 1.4}$
& $\mathbf{95.3 \pm 1.0}$
& $\underline{0.478}$
\\

& PointNetGPD~\cite{liang2019pointnetgpd}
& $\times$
& $\underline{99.9 \pm 0.1}$
& $\underline{97.9 \pm 0.7}$
& $\underline{92.2 \pm 1.3}$
& $\mathbf{92.9 \pm 2.6}$
& $\mathbf{91.0 \pm 1.4}$
& $\mathbf{92.8 \pm 1.5}$
& $\underline{94.7 \pm 1.9}$
& $3.308$
\\

& GraspGen Disc.~\cite{murali2025graspgen}
& $\checkmark$
& $98.9 \pm 0.9$
& $94.1 \pm 1.2$
& $91.6 \pm 1.8$
& $81.4 \pm 3.5$
& $71.1 \pm 2.3$
& $69.2 \pm 3.5$
& $81.2 \pm 3.6$
& $\mathbf{0.254}$
\\

\midrule

\multirow{3}{*}{Samples from GraspGen Gen.$^{*}$~\cite{murali2025graspgen}}
& EquiGQNet
& $\checkmark$
& $\mathbf{97.8 \pm 2.1}$
& $\mathbf{97.3 \pm 1.0}$
& $\mathbf{93.6 \pm 1.1}$
& $\underline{84.8 \pm 4.1}$
& $\underline{83.0 \pm 1.9}$
& $\mathbf{87.2 \pm 2.3}$
& $\mathbf{94.8 \pm 1.9}$
& $\mathbf{0.349}$
\\

& PointNetGPD~\cite{liang2019pointnetgpd}
& $\times$
& $\underline{97.5 \pm 2.2}$
& $94.7 \pm 1.3$
& $88.6 \pm 1.6$
& $\mathbf{86.7 \pm 3.8}$
& $\mathbf{85.0 \pm 1.8}$
& $\underline{87.0 \pm 1.9}$
& $\underline{94.2 \pm 1.9}$
& $0.561$
\\

& GraspGen Disc.~\cite{murali2025graspgen}
& $\checkmark$
& $\underline{97.5 \pm 2.3}$
& $\underline{97.1 \pm 0.9}$
& $\underline{92.2 \pm 1.2}$
& $80.3 \pm 4.2$
& $75.7 \pm 2.2$
& $77.4 \pm 2.7$
& $80.1 \pm 4.6$
& $\underline{0.357}$
\\

\midrule

\multirow{2}{*}{Scene-anchored grasp methods}
& EdgeGrasp$^{*}$~\cite{huang2023edgegrasp}
& $\checkmark$
& $\underline{89.9 \pm 3.5}$
& $\underline{76.1 \pm 2.2}$
& $\underline{73.4 \pm 2.5}$
& $\underline{62.5 \pm 5.4}$
& $\underline{55.6 \pm 2.5}$
& $\underline{48.4 \pm 3.6}$
& $\underline{69.5 \pm 4.9}$
& $\mathbf{0.466}$
\\

& OrbitGrasp$^{*}$~\cite{hu2024orbitgrasp}
& $\checkmark$
& $\mathbf{97.4 \pm 1.7}$
& $\mathbf{94.9 \pm 1.2}$
& $\mathbf{89.6 \pm 1.5}$
& $\mathbf{81.6 \pm 3.1}$
& $\mathbf{76.3 \pm 1.9}$
& $\mathbf{75.8 \pm 2.9}$
& $\mathbf{78.0 \pm 3.9}$
& $\underline{2.331}$
\\

\bottomrule
\multicolumn{11}{l}{
\footnotesize
$^{*}$ denotes the use of a pretrained model.
}
\\[-1pt]
\multicolumn{11}{l}{
\footnotesize
\textbf{Bold} and \underline{underlined} values indicate the best and second-best results within each search strategy, respectively.
}
\\[-1pt]
\multicolumn{11}{l}{
\footnotesize
For GSR and DR, $\pm$ denotes the cluster bootstrap standard error (20{,}000 resamples), with objects as clusters for isolated-object evaluation and scenes for cluttered evaluation.
}
\end{tabular}%
}
\end{minipage}

\par\vspace{2mm}

\begin{minipage}{\textwidth}
\centering
\captionof{table}{Ablation study of EquiGQNet}
\label{tab:ablation}
\resizebox{\textwidth}{!}{%
\begin{tabular}{ccccc ccc ccc c c}
\toprule
\multirow{3}{*}{Method}
&
\multicolumn{2}{c}{Fusion Method}
&
\multirow{3}{*}{Encoder}
&
\multirow{3}{*}{\makecell{Shared\\Scene\\Encoding}}
&
\multicolumn{3}{c}{IPA-3D1K}
&
\multicolumn{3}{c}{Thingi10K}
&
\multicolumn{1}{c}{Dex-Net Adv.}
&
\multirow{3}{*}{\makecell{Planning\\Time (s)\\($N_q=5$)}}
\\
\cmidrule(lr){2-3}
\cmidrule(lr){6-8}
\cmidrule(lr){9-11}
\cmidrule(lr){12-12}
&
Rotation
&
Translation
&
&
&
\multicolumn{1}{c}{\makecell{Isolated\\Object}}
&
\multicolumn{2}{c}{Cluttered}
&
\multicolumn{1}{c}{\makecell{Isolated\\Object}}
&
\multicolumn{2}{c}{Cluttered}
&
\multicolumn{1}{c}{\makecell{Isolated\\Object}}
&
\\
\cmidrule(lr){6-6}
\cmidrule(lr){7-8}
\cmidrule(lr){9-9}
\cmidrule(lr){10-11}
\cmidrule(lr){12-12}
&
&
&
&
&
GSR (\%)
&
GSR (\%)
&
DR (\%)
&
GSR (\%)
&
GSR (\%)
&
DR (\%)
&
GSR (\%)
&
\\

\midrule

(a)
& global
& global
& DGCNN
& $\checkmark$
& $98.2 \pm 0.8$
& $91.4 \pm 1.2$
& $87.4 \pm 2.0$
& $74.1 \pm 4.2$
& $60.1 \pm 2.3$
& $57.2 \pm 3.5$
& $66.6 \pm 4.4$
& $\mathbf{0.211}$
\\

(b)
& canon.
& global
& DGCNN
& $\times$
& $98.3 \pm 0.7$
& $96.8 \pm 0.9$
& $92.0 \pm 1.1$
& $75.8 \pm 4.4$
& $63.2 \pm 2.1$
& $63.8 \pm 3.1$
& $89.0 \pm 2.7$
& $7.199$
\\

(c)
& canon.
& MAF
& DGCNN
& $\times$
& $\mathbf{99.8 \pm 0.3}$
& $\underline{99.4 \pm 0.4}$
& $\mathbf{96.2 \pm 0.8}$
& $\underline{88.7 \pm 2.9}$
& $\mathbf{85.6 \pm 1.8}$
& $\mathbf{89.0 \pm 2.0}$
& $\underline{97.0 \pm 1.0}$
& $7.222$
\\

(d)
& canon.
& canon.
& DGCNN
& $\times$
& $\underline{99.7 \pm 0.3}$
& $\mathbf{99.6 \pm 0.3}$
& $\underline{95.0 \pm 0.9}$
& $\mathbf{89.1 \pm 2.8}$
& $\underline{84.8 \pm 1.7}$
& $\underline{87.2 \pm 2.1}$
& $\mathbf{98.5 \pm 0.8}$
& $\underline{6.998}$
\\

\midrule

(e)
& canon.
& MAF
& VN-DGCNN
& $\checkmark$
& $\underline{99.7 \pm 0.3}$
& $\underline{99.2 \pm 0.4}$
& $\mathbf{96.0 \pm 0.9}$
& $\underline{85.0 \pm 3.3}$
& $\underline{80.2 \pm 1.7}$
& $\underline{85.0 \pm 2.3}$
& $\underline{95.2 \pm 1.3}$
& $\underline{0.481}$
\\

EquiGQNet
& canon.
& MAF
& GVN-DGCNN
& $\checkmark$
& $\mathbf{100.0 \pm 0.1}$
& $\mathbf{99.8 \pm 0.2}$
& $\underline{95.2 \pm 0.9}$
& $\mathbf{89.5 \pm 2.8}$
& $\mathbf{86.0 \pm 1.6}$
& $\mathbf{92.2 \pm 1.4}$
& $\mathbf{95.3 \pm 1.0}$
& $\mathbf{0.478}$
\\

\midrule

(f)
& global
& global
& GVN-DGCNN
& $\checkmark$
& $70.2 \pm 1.9$
& $42.0 \pm 3.1$
& $29.2 \pm 3.3$
& $45.3 \pm 3.3$
& $24.4 \pm 2.5$
& $14.6 \pm 2.2$
& $34.2 \pm 3.0$
& $\mathbf{0.227}$
\\

(g)
& canon.
& global
& GVN-DGCNN
& $\checkmark$
& $\underline{89.3 \pm 2.5}$
& $\underline{77.7 \pm 2.2}$
& $\underline{76.6 \pm 3.2}$
& $\underline{61.8 \pm 5.0}$
& $\underline{37.7 \pm 2.9}$
& $\underline{26.4 \pm 3.2}$
& $\underline{60.8 \pm 4.0}$
& $\underline{0.230}$
\\

(h)
& MAF
& MAF
& DGCNN
& $\checkmark$
& $\mathbf{99.6 \pm 0.4}$
& $\mathbf{97.9 \pm 0.7}$
& $\mathbf{94.2 \pm 1.2}$
& $\mathbf{84.2 \pm 3.4}$
& $\mathbf{72.5 \pm 2.0}$
& $\mathbf{75.6 \pm 3.0}$
& $\mathbf{93.8 \pm 2.0}$
& $0.498$
\\

\bottomrule
\multicolumn{13}{l}{
\footnotesize
\textbf{Bold} and \underline{underlined} values indicate the best and second-best results
within each comparison sector, respectively.
}
\\[-1pt]
\multicolumn{13}{l}{
\footnotesize
For GSR and DR, $\pm$ denotes the cluster bootstrap standard error (20{,}000 resamples), with objects as clusters for isolated-object evaluation and scenes for cluttered evaluation.
}
\end{tabular}%
}
\end{minipage}

\par\vspace{1mm}


\includegraphics[width=\textwidth]
{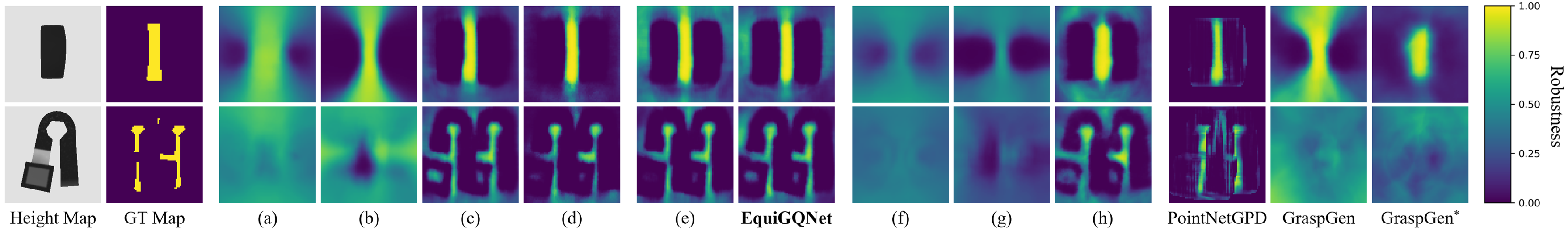}

\caption{
Grasp robustness maps for an IPA-3D1K object (top) and a Dex-Net adversarial object (bottom).
For a fixed grasp orientation and height, grasp quality is evaluated over a dense grid of planar translations.
Columns show the object height map, ground-truth (GT) robustness, predictions from ablation variants (a)--(h) in Table~\ref{tab:ablation}, PointNetGPD~\cite{liang2019pointnetgpd}, our retrained GraspGen discriminator, and the official pretrained GraspGen discriminator (GraspGen$^*$)~\cite{murali2025graspgen}.
Methods that incorporate grasp information before global aggregation produce more spatially localized quality landscapes, particularly for the challenging adversarial geometry.
}
\label{fig:grasp_robustness_map}
\vspace{-2mm}
\end{figure*}

%% file: tables/validation_accuracy.tex

\begin{table}[t]
\centering
\caption{Validation performance of grasp-quality predictors.}
\label{tab:validation_accuracy}

\setlength{\tabcolsep}{5pt}
\renewcommand{\arraystretch}{1.05}

\begin{tabular}{c c c c}
\toprule
Method & Acc. (\%) & AP (\%) & AUROC (\%) \\
\midrule

(a) & 86.78 & 71.57 & 91.60 \\
(b) & 89.03 & 78.82 & 93.93 \\
(c) & \underline{91.37} & \underline{85.98} & \underline{96.24} \\
(d) & \textbf{91.82} & \textbf{87.12} & \textbf{96.61} \\

\midrule

(e) & \textbf{91.23} & \textbf{85.79} & \textbf{96.27} \\
EquiGQNet & \textbf{91.23} & \underline{85.70} & \textbf{96.27} \\

PointNetGPD~\cite{liang2019pointnetgpd}
& 91.09 & 84.17 & 95.99 \\

GraspGen Disc.~\cite{murali2025graspgen}
& 88.35 & 76.89 & 93.32 \\

\midrule

(f) & 84.56 & 62.56 & 88.53 \\
(g) & \underline{85.89} & \underline{67.75} & \underline{90.09} \\
(h) & \textbf{90.39} & \textbf{83.12} & \textbf{95.38} \\

\bottomrule
\end{tabular}

\vspace{2pt}
\begin{minipage}{0.98\columnwidth}
\footnotesize
Methods (a)--(h) are defined in Table~\ref{tab:ablation}.
Ground-truth robustness is binarized at 0.5; accuracy additionally thresholds the predicted score at 0.5, whereas AP and AUROC are computed from the continuous predicted scores.
\\
\textbf{Bold} and \underline{underlined} values indicate the best and second-best results within each comparison sector, respectively.
\end{minipage}

\end{table}

%% file: tables/physical_experiments.tex
\begin{table}[t]
\centering
\caption{Experimental results of different grasp evaluation methods.}
\label{tab:real_experiment}

\footnotesize
\setlength{\tabcolsep}{1.5pt}
\renewcommand{\arraystretch}{1.05}

\begin{tabular}{@{}l ccc ccc@{}}
\toprule

\multirow{2}{*}{Method}
&
\multicolumn{3}{c}{Household Objects}
&
\multicolumn{3}{c}{Dex-Net Adv.}
\\

\cmidrule(lr){2-4}
\cmidrule(lr){5-7}

&
GSR (\%)
&
DR (\%)
&
MPPH
&
GSR (\%)
&
DR (\%)
&
MPPH
\\

\midrule

EquiGQNet
&
\makecell{\textbf{95.2}\\[-0.5pt]{\scriptsize (120/126)}}
&
\makecell{\textbf{100}\\[-0.5pt]{\scriptsize (120/120)}}
&
\textbf{230}
&
\makecell{\underline{93.8}\\[-0.5pt]{\scriptsize (120/128)}}
&
\makecell{\textbf{100}\\[-0.5pt]{\scriptsize (120/120)}}
&
\textbf{206}
\\

\addlinespace[1.2pt]

PointNetGPD~\cite{liang2019pointnetgpd}
&
\makecell{\underline{94.8}\\[-0.5pt]{\scriptsize (109/115)}}
&
\makecell{\underline{90.8}\\[-0.5pt]{\scriptsize (109/120)}}
&
153
&
\makecell{\textbf{94.4}\\[-0.5pt]{\scriptsize (118/125)}}
&
\makecell{\underline{98.3}\\[-0.5pt]{\scriptsize (118/120)}}
&
\underline{135}
\\

\addlinespace[1.2pt]

GraspGen Disc.~\cite{murali2025graspgen}
&
\makecell{77.1\\[-0.5pt]{\scriptsize (84/109)}}
&
\makecell{70.0\\[-0.5pt]{\scriptsize (84/120)}}
&
\underline{170}
&
\makecell{66.7\\[-0.5pt]{\scriptsize (56/84)}}
&
\makecell{46.7\\[-0.5pt]{\scriptsize (56/120)}}
&
125
\\

\bottomrule
\end{tabular}

\vspace{2pt}
\begin{minipage}{0.98\columnwidth}
\footnotesize
GSR = successful grasps/attempts; DR = removed/initial objects.
\\
\textbf{Bold} and \underline{underlined} values indicate the best and second-best results for each metric within each object set, respectively.
\end{minipage}

\end{table}

%% file: conclusion.tex
\section{Conclusion}
\label{sec:conclusion}

We presented \textbf{EquiGQNet}, an efficient grasp quality evaluator that reuses $\mathrm{SO}(3)$-equivariant scene features across arbitrary 6-DoF poses while retaining local geometry relative to each grasp through explicit orientation alignment and pre-aggregation translation fusion (MAF).
This avoids the repeated scene encoding required by candidate-wise early fusion and substantially outperforms global late fusion on challenging geometries, producing more localized quality landscapes that matter when valid grasps occupy narrow regions, while achieving the highest real-world throughput at comparable success rates.
Two limitations remain. Despite shared encoding, EquiGQNet remains about $1.9\times$ slower in CEM planning than pure global late fusion (0.478~s vs.\ 0.254~s at five local scene queries), motivating lighter equivariant backbones and fusion modules.
In addition, while EquiGQNet supports arbitrary 6-DoF grasp evaluation, our experiments focus on tabletop settings with a $45^\circ$ approach-angle constraint.
Extending the evaluation to environments requiring side approaches, such as shelves or bins, remains future work.